\documentclass[lettersize,journal]{IEEEtran}
\usepackage{amsmath}
\usepackage{amsfonts}
\usepackage{algorithmic}
\usepackage{algorithm}
\usepackage{array}
\usepackage[caption=false,font=normalsize,labelfont=sf,textfont=sf]{subfig}
\usepackage{textcomp}
\usepackage{stfloats}
\usepackage{url}
\usepackage{verbatim}
\usepackage{graphicx}
\usepackage{cite}
\usepackage{booktabs}
\usepackage{xcolor}
\usepackage[citecolor=blue, colorlinks]{hyperref}

\begin{document}

\title{PointMCD: Boosting Deep Point Cloud Encoders \\ via Multi-view Cross-modal Distillation \\ for 3D Shape Recognition}

\author{Qijian Zhang, Junhui Hou, \textit{Senior Member}, \textit{IEEE}, and Yue Qian
	\IEEEcompsocitemizethanks{
			\IEEEcompsocthanksitem This project was supported by the Hong Kong Research
		Grants Council under Grants 11202320 and 11218121. \textit{Corresponding author: Junhui Hou}
		\IEEEcompsocthanksitem All authors are with the Department of Computer Science, City University of Hong Kong, Hong Kong SAR. Email: qijizhang3-c@my.cityu.edu.hk; jh.hou@cityu.edu.hk; yueqian4-c@my.cityu.edu.hk;}
}

\maketitle

\begin{abstract}
	As two fundamental representation modalities of 3D objects, 3D point clouds and multi-view 2D images record shape information from different domains of geometric structures and visual appearances. In the current deep learning era, remarkable progress in processing such two data modalities has been achieved through respectively customizing compatible 3D and 2D network architectures. However, unlike multi-view image-based 2D visual modeling paradigms, which have shown leading performance in several common 3D shape recognition benchmarks, point cloud-based 3D geometric modeling paradigms are still highly limited by insufficient learning capacity due to the difficulty of extracting discriminative features from irregular geometric signals. In this paper, we explore the possibility of boosting deep 3D point cloud encoders by transferring visual knowledge extracted from deep 2D image encoders under a standard teacher-student distillation workflow. Generally, we propose PointMCD, a unified multi-view cross-modal distillation architecture, including a pretrained deep image encoder as the teacher and a deep point encoder as the student. To perform heterogeneous feature alignment between 2D visual and 3D geometric domains, we further investigate visibility-aware feature projection (VAFP), by which point-wise embeddings are reasonably aggregated into view-specific geometric descriptors. By pair-wisely aligning multi-view visual and geometric descriptors, we can obtain more powerful deep point encoders without exhausting and complicated network modification. Experiments on 3D shape classification, part segmentation, and unsupervised learning strongly validate the effectiveness of our method. \textit{The code and data will be publicly available at \url{https://github.com/keeganhk/PointMCD}}.
\end{abstract}

\begin{IEEEkeywords}
	3D point cloud, multi-view images, cross-modal, knowledge distillation, 3D shape recognition
\end{IEEEkeywords}

\section{Introduction} \label{sec:intro}

\IEEEPARstart{D}{riven} by the recent advancements and popularization of 3D acquisition and perception technologies, 3D shape recognition has been attracting increasingly growing attention in both industry and academia. Generally, different from its scene-level counterparts such as place recognition \cite{uy2018pointnetvlad}, indoor \cite{hua2016scenenn,armeni20163d,dai2017scannet} or outdoor \cite{hackel2017semantic3d,roynard2018paris,tan2020toronto} semantic segmentation, and LiDAR (i.e., light detection and ranging) driving environment object detection \cite{geiger2012we,sun2020scalability,caesar2020nuscenes}, shape recognition tasks focus on object-level models and thus require a more fine-grained understanding of 3D geometry, which pose special challenges and differentiated emphases in this field.

In contrast to 2D visual signals that are typically represented by images captured from ordinary optical cameras, 3D shape information can be recorded by multiple alternative modalities generated by rich varieties of 3D sensors/scanners, which are widely deployed in different application scenarios. In the deep learning era, depending on different representation modalities of 3D objects, mainstream learning pipelines can be classified into voxel-based \cite{maturana2015voxnet,wu20153d,riegler2017octnet,wang2017cnn,klokov2017escape}, image-based \cite{su2015multi,feng2018gvcnn,yu2018multi,kanezaki2018rotationnet,yang2019learning,esteves2019equivariant,wei2020view,hamdi2021mvtn}, and point-based \cite{qi2017pointnet,qi2017pointnet++,xu2018spidercnn,li2018pointcnn,liu2019relation,thomas2019kpconv,wang2019dynamic,zhao2021point,xiang2021walk,ma2022rethinking} processing paradigms. Generally, voxel-based models employ regular volumetric grids to describe the spatial occupancy status of irregular geometric structures, such that standard 3D convolutional neural network (CNN) architectures can be naturally and seamlessly introduced. Unfortunately, due to the cubic growth of computational complexity and memory footprint, these methods are not suitable for dealing with high-resolution volumes with fine details, despite the exploration of more complicated adaptive and hierarchical spatial indexing strategies \cite{riegler2017octnet,wang2017cnn,klokov2017escape}. Instead of faithfully recording 3D spatial structures in the original geometric space, image-based models focus on visual appearances of object surfaces by rendering a collection of multi-view 2D images from different viewpoints. Benefiting from the maturity of powerful 2D modeling architectures \cite{simonyan2014very,he2016deep,long2015fully,ronneberger2015u} and the availability of large-scale annotated image databases \cite{krizhevsky2012imagenet,lin2014microsoft,cordts2016cityscapes}, such multi-view learning paradigm has demonstrated dominating performance in common shape recognition benchmarks \cite{chang2015shapenet,wu20153d} for diverse tasks of classification, retrieval, and pose estimation. Meanwhile, there also exist several aspects of limitations such as ignorance of interior spatial structures, difficulty in topology analysis, and inevitable loss of detailed surface textures.

More recently, point-based models that can directly work on unstructured point clouds have gained popularity. As the most straightforward representation modality for geometric signals, point clouds serve as raw outputs of many 3D data acquisition systems and faithful records of the original spatial information. However, different from images/voxels defined on regular grid domains, point cloud data are characterized by irregularity and unorderedness, causing mush difficulty in designing expressive feature extraction operators with sufficient modeling capacity. Despite the proliferation of various fancy designs of deep set architectures in previous works, there still exists a lot of room for further performance boost. In practice, there is no unified modeling paradigm for 3D shape understanding, as each data modality and model architecture uniquely have different merits and limitations.

In fact, image-driven 2D visual modeling and point-driven 3D geometric modeling have a degree of complementarity. On one hand, image-based models built upon well-developed 2D CNNs and sufficiently-large image datasets outperform point-based models in terms of network capacity and generalization ability. On the other hand, since 3D point clouds intrinsically contain more complete geometric information than multi-view 2D image renderings, we may expect that point-based models indicate a more promising direction with greater potential and convenience.

\begin{figure*}[t]
	\centering
	\includegraphics[width=0.90\linewidth]{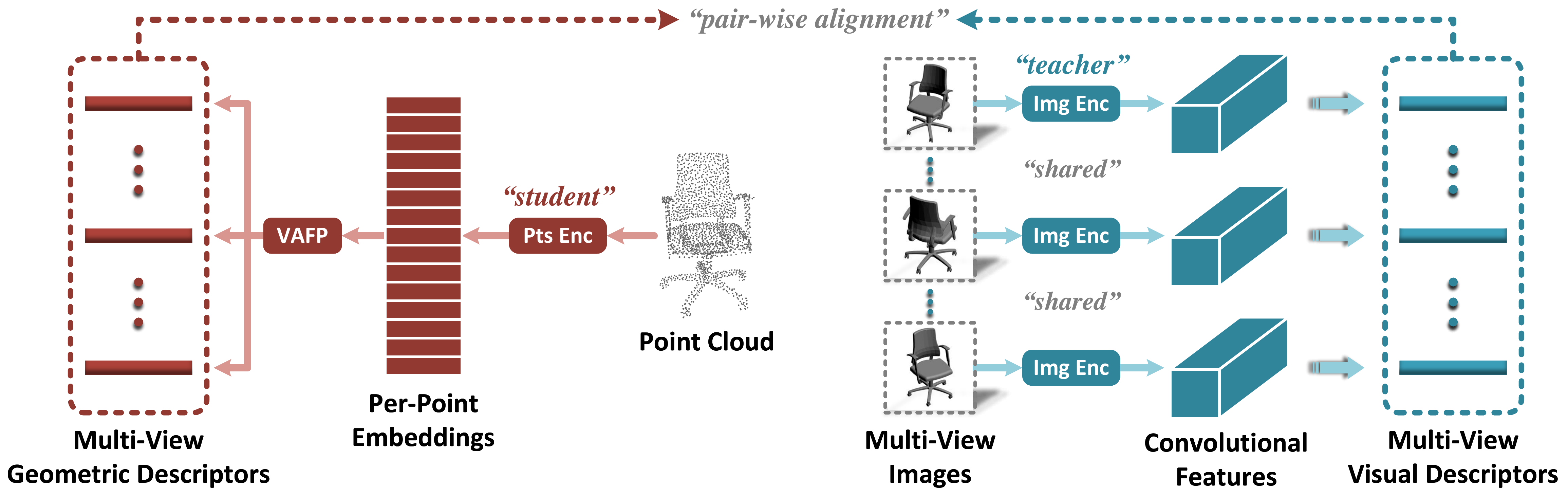}
	\caption{Flowchart of the proposed PointMCD for multi-view cross-modal distillation. In the (right) teacher branch, a collection of multi-view images are fed into shared deep 2D CNNs to generate convolutional feature maps, which are further converted into vectorized multi-view visual descriptors. In the (left) student branch, the target deep 3D point encoder consumes the input point cloud and extracts high-dimensional point-wise embeddings, which are further aggregated via the proposed novel visibility-aware feature projection (VAFP) mechanism into vectorized multi-view geometric descriptors. Intuitively, VAFP selects points that are visible from a certain viewpoint, and then collects feature embeddings of the selected visible points for further aggregation. Thus, by pair-wisely aligning multi-view visual and geometric descriptors, the student model can be guided to learn more discriminative point features.}
	\label{fig:overall-workflow}
\end{figure*}

In this paper, instead of investigating new backbone network architectures for point feature extraction, we seek to construct a unified processing pipeline that can effectively boost existing mainstream deep point encoders to learn more discriminative features. Technically, we propose PointMCD that implements multi-view cross-modal distillation by transferring knowledge from a pretrained deep 2D image encoder as the teacher to a target deep 3D point encoder as the student, which is a highly non-trivial problem due to significant domain gaps in terms of both data modality and model structure. Fig. \ref{fig:overall-workflow} presents the block diagram of the overall workflow. In the teacher branch, we feed a collection of multi-view images rendered from a pre-defined set of camera poses into a  pretrained deep 2D image encoder, from which we obtain vectorized multi-view visual descriptors serving as teacher knowledge to be transferred. In the student branch, the target deep 3D point encoder consumes the corresponding point cloud model and produces point-wise embeddings. After that, to facilitate multi-view heterogeneous feature alignment between visual and geometric domains, we introduce a novel  visibility-aware feature projectio (VAFP) mechanism that aggregates vectorized multi-view geometric descriptors on the basis of view-specific per-point visibility computed under the same set of pre-defined camera poses. Finally, the distillation process can be achieved by pair-wisely imposing alignment constraints on the produced multi-view visual and geometric descriptors.

To verify the \textit{effectiveness} and \textit{universality} of our proposed PointMCD, we select three classic deep point encoders \cite{qi2017pointnet,qi2017pointnet++,wang2019dynamic} as well as a more recent state-of-the-art framework named CurveNet \cite{xiang2021walk} as the baseline student models, and conduct experiments on three common benchmark tasks, including 3D shape classification, part segmentation, and unsupervised learning, where we can observe stable and encouraging performance gains.

In summary, our main contributions are three-fold:

\begin{itemize}
	\item we propose a unified multi-view cross-modal distillation framework, namely PointMCD, which aims at boosting deep 3D point cloud encoders without additional efforts on model structure modification;
	\item to facilitate multi-view heterogeneous (visual-geometric) feature alignment, we develop a novel VAFP mechanism, which reasonably aggregates point-wise embeddings into view-specific geometric descriptors on the basis of physical visibility property; and
	\item we achieve a highly encouraging performance boost on a rich variety of baseline models and downstream application scenarios.
\end{itemize}

Overall, the main technical novelty of this work lies in that we skillfully integrate two heterogeneous 3D data representation structures -- 2D multi-view images and 3D point clouds -- into a unified cross-modal knowledge distillation pipeline via customizing view-specific visual-geometric feature alignment. Our exploration implies a generic and promising direction for enhancing the learning capacity of point-based 3D geometric modeling architectures.

The remainder of this paper is organized as follows. In Sec. \ref{sec:rel-work}, we detailedly review three aspects of closely-related works. In Sec. \ref{sec:proposed-method}, we introduce the overall workflow of the proposed PointMCD learning framework. We report experimental results in Sec. \ref{sec:experiments}, including evaluations on three benchmark tasks of 3D shape recognition, comparisons with baseline models and other state-of-the-art methods, and carefully designed ablation studies. In Sec. \ref{sec:discussion}, we re-emphasize the design philosophy of our specific technical implementations and further discuss new insights revealed by this work. In the end, we summarize our paper in Sec. \ref{sec:conclusion}.

\section{Related Work} \label{sec:rel-work}

In this section, we start by reviewing both multi-view image-based visual modeling paradigms in Sec. \ref{sec:II.A} and point-based geometric modeling paradigms in Sec. \ref{sec:II.B}, which serve as the two main ingredients in our proposed cross-modal processing pipeline. In \ref{sec:II.C}, we discuss the current research progress in knowledge distillation, especially for the most closely related works with focuses on big-gap cross-modal transfer between 2D and 3D domains.

\subsection{Multi-View 2D CNNs for 3D Shape Recognition} \label{sec:II.A}

As a natural extension of 2D deep learning frameworks for image processing \cite{wang2022mixture,si2022hybrid}, multi-view 3D shape modeling is typically built upon various variants of multi-input 2D CNNs.

Pioneered by MVCNN \cite{su2015multi} that separately consumes multi-view images rendered from multiple virtual camera poses and generates vectorized global shape signature through cross-view max-pooling, many follow-up studies are devoted to designing more advanced learning strategies of inter-view interaction and viewpoint selection. GVCNN \cite{feng2018gvcnn} proposes a three-level hierarchical correlation modeling framework, which adaptively groups multi-view feature embeddings into separate clusters. MHBN \cite{yu2018multi} and RelationNet \cite{yang2019learning} further exploit patch-level interaction to enrich inter-view relationships. RotationNet \cite{kanezaki2018rotationnet} treats viewpoint indices as learnable latent variables and tends to jointly estimate object poses and semantic categories. EMV \cite{esteves2019equivariant} presents a group convolution approach that operates on a discrete subgroup of rotation groups, which enable to extract rotation-equivalent shape descriptors. View-GCN \cite{wei2020view} regards viewpoints as graph nodes to construct a directed view graph, on which graph convolution can be applied to learn inter-view relations. MVTN \cite{hamdi2021mvtn} introduces the differentiable rendering techniques to implement adaptive regression of optimal camera poses in an end-to-end trainable manner.

Generally, image-based methods have demonstrated leading performance in common shape recognition benchmarks \cite{chang2015shapenet,wu20153d} for tasks like classification, retrieval, and pose estimation. Nevertheless, adapting such image-based learning paradigm to topology analysis and some fine-grained prediction scenarios (e.g., segmentation, normal estimation) is known to be highly non-trivial and cumbersome. Besides, interior spatial structures are totally ignored during surface rendering.

\subsection{Deep Set Architectures for 3D Point Cloud Processing} \label{sec:II.B}

In real-world applications, point cloud data play a key role in numerous low-level and high-level task scenarios \cite{de2018graph,li2020efficient,zhang2020pointhop,liu2020semantic,valsesia2020learning,chen2020hapgn,guarda2020constant,qiu2021geometric}. Different from the maturity of 2D deep learning frameworks for image processing, the exploration of deep set architectures for 3D point cloud modeling is still at its fast-growing stage.

Pioneered by PointNet \cite{qi2017pointnet} that adopts point-wisely shared multi-layer perceptrons (MLPs) for permutation-invariant feature embedding, deep set architectures that directly operate on unstructured 3D point clouds have rapidly gained popularity in the geometry processing community. Inheriting the successful design experience of 2D CNNs, PointNet++ \cite{qi2017pointnet++} incorporates local neighborhood aggregation and further adopts hierarchical feature abstraction by applying progressive downsampling. A rich variety of highly-specialized point cloud feature aggregation operators have been developed in later studies. Typically, \cite{hua2018pointwise,li2018pointcnn,verma2018feastnet,thomas2019kpconv,xu2021paconv} mimic standard convolutions by learning adaptive weights to implement kernel matching. Following the design philosophy of specializing point convolutions, \cite{li2018so,liu2019relation,wu2019pointconv,zhang2019shellnet} are devoted to designing various more complicated point feature aggregation strategies to enhance network capacity. Another classic work DGCNN \cite{wang2019dynamic} investigates a graph-based dynamic feature updating mechanism, which can capture global contextual information in a flexible manner. \cite{yang2019modeling,nezhadarya2020adaptive,yan2020pointasnl} explore learning-based, instead of heuristic, subset selection techniques to adaptively preserve the most informative points after downsampling. Driven by the latest trend in 2D vision and language community, transformer-style architectures \cite{zhao2021point,hui2021pyramid,mazur2021cloud,krivokuca2021compression,jiayao2022real} are also adapted for point cloud modeling, and their corresponding self-supervised pre-training schemes \cite{yu2022point,pang2022masked} show impressive performances in various downstream tasks.

Thanks to the recent academic attention, we have witnessed continuous progress in learning deep features directly from 3D point clouds. However, point-based methods still suffer from insufficient modeling capacity and relatively sub-optimal task performance. Moreover, since 3D labeling can be much more difficult and expensive than 2D image annotation, existing 3D shape repositories are of small scale and limited richness.

\subsection{Knowledge Transfer between 2D and 3D Modalities} \label{sec:II.C}

Ever since Hinton \textit{et al}. \cite{hinton2015distilling} firstly proposed the processing paradigm of knowledge distillation, there have been numerous academic explorations of diverse technical variants and wide deployment in real applications. However, as analyzed in \cite{gou2021knowledge}, transferring knowledge across multi-modalities still remains a highly challenging problem, especially when the domain gap is big (e.g., lacking of paired samples), and such practice turns to be even more non-trivial when conducted between 2D and 3D domains with an obvious dimensionality gap.

Typically, xMUDA \cite{jaritz2020xmuda} proposes to achieve unsupervised domain adaptation from the 2D source domain of single-view camera images to the 3D target domain of LiDAR point clouds by aligning outputs of 2D and 3D branches according to pixel-point correspondences. PPKT \cite{liu2021learning} constructs a 3D pretraining framework to leverage 2D pretrained knowledge by applying the contrastive learning strategy on positive and negative pixel-point pairs. In an opposite transferring direction, Pri3D \cite{hou2021pri3d} explores the potential of 3D-guided contrastive pretraining for boosting 2D perception capability. In addition to maintaining feature consistency between paired 2D pixels and 3D points, this work also aims at learning invariant pixel-wise descriptors across the image scans captured from different viewpoints. A more flexible 3D-to-2D distillation framework can be found in \cite{liu20213d}, which aligns statistical distributions of convolutional features from 2D and 3D CNNs through specialized dimension normalization techniques. To get rid of the dependence on fine-grained correspondences between 2D and 3D modalities that are usually expensive to acquire, this work further explores a semantics-aware adversarial training scheme, which enables to tackle unpaired 2D images and 3D volumetric grids.

Conclusively, existing studies on cross-modal (2D and 3D) knowledge transfer mainly focus on scene-level parsing, owing to the convenience of computing fine-grained correspondences (i.e., pixel-point/voxel pairs) from the corresponding benchmark datasets \cite{dai2017scannet,caesar2020nuscenes,behley2019semantickitti} with synchronized and calibrated camera and LiDAR device configuration, and actually we also need to shift more attention to shape recognition (object-level) scenarios. More importantly, heterogeneous feature alignment, which plays a key role in cross-modal distillation, still needs to be implemented in a more effective way that better boosts the distilled model.

\section{Proposed Method} \label{sec:proposed-method}

This section introduces the overall working mechanism and specific technical implementations of the proposed PointMCD. To facilitate understanding, we begin with an overview of our framework in Sec. \ref{sec:III.A}. Then we present general formulations of deep image encoders and deep point encoders respectively in Sec. \ref{sec:III.B} and Sec. \ref{sec:III.C}, based on which we construct a unified cross-modal distillation pipeline driven by multi-view visibility-aware feature alignment in Sec. \ref{sec:III.D}. In the end, we summarize the overall loss function and training strategy in Sec. \ref{sec:III.E}.

\subsection{Problem Overview} \label{sec:III.A}

We aim to transfer cross-modal knowledge from pretrained deep image encoders that extract 2D visual features to deep point encoders that extract 3D geometric embeddings.

Formally, we consider a given 3D point cloud $\mathcal{P} \in \mathbb{R}^{N \times 3}$ containing $N$ spatial points and a collection of 2D multi-view images $\{\mathcal{I}_k \in \mathbb{R}^{H \times W \times 3}\}_{k=1}^{K}$ capturing view-specific visual appearances of the original 3D shape geometry when rendered from $K$ pre-defined camera positions $\{\mathbf{c}_k=(\theta^{az}_k, \theta^{el}_k, \mu)\}_{k=1}^{K}$, where $\theta^{az}_k$ and $\theta^{el}_k$ denote azimuth and elevation angles relative to the object centroid at viewpoint $\mathbf{c}_k$ and $\mu$ is specified as a constant observation distance.

Following the standard teacher-student architectural design, the proposed PointMCD is composed of a 2D teacher branch $\mathcal{B}_t = \{ \mathcal{M}_t, \mathcal{H}_t \}$ and a 3D student branch $\mathcal{B}_s = \{ \mathcal{M}_s, \mathcal{H}_s \}$, in which $\mathcal{M}_t$ and $\mathcal{M}_s$ respectively represent the pretrained deep image encoder to be transferred and the deep point encoder to be distilled. We perform distillation by aligning geometric embeddings learned by $\mathcal{M}_s$ with visual features exported from $\mathcal{M}_t$. The subsequent task-specific head networks are denoted as $\mathcal{H}_t$ and $\mathcal{H}_s$, which are excluded from the actual distillation process. Functionally, we emphasize that PointMCD serves as a universal processing pipeline that is compatible with generic deep set architectures.

\textbf{Remark}. In particular, it is worth reminding that our method should be differentiated from multi-modal fusion frameworks that consume data of multi-modalities as inputs during both training and inference stages. In other words, the core concern of this work lies in information transfer from source domain to target domain, instead of multi-source feature fusion and interaction (e.g., \cite{xu2021rpvnet}).

\subsection{Teacher Network for 2D Image Modeling} \label{sec:III.B}

Driven by the maturity of deep convolutional architectures, we resort to powerful 2D CNNs in the visual modeling branch for image feature extraction. On one hand, the proliferation of numerous off-the-shelf deep architectures \cite{simonyan2014very,he2016deep,szegedy2016rethinking,sandler2018mobilenetv2} allows us to conveniently deploy 2D image backbone encoders without additional efforts on model design. On the other hand, benefiting from the common practice of large-scale pretraining (\textit{e.g.,} on ImageNet \cite{krizhevsky2012imagenet}), the resulting deep feature extractors demonstrate satisfactory generalization ability when fine-tuned on downstream visual recognition application scenarios. These excellent properties make 2D CNNs an optimal choice of the teacher model of the deep image encoder $\mathcal{M}_t$.

Formally, the teacher encoder consumes multi-view images in parallel as inputs and correspondingly produce a collection of high-dimensional 2D convolutional feature maps. A general abstraction of the teacher model behaviors is formulated as
\begin{equation} \label{general-teacher-formulation}
	\{\mathcal{V}_k\}_{k=1}^{K} = \mathcal{M}_t\left(\{\mathcal{I}_k\}_{k=1}^{K}\right),
\end{equation}
\noindent where $\mathcal{V}_k \in \mathbb{R}^{H^{\prime} \times W^{\prime} \times C_t}$ with subscript of $k$ represents a view-specific 2D convolutional feature map extracted from image $\mathcal{I}_k$. This corresponds to partial encoding of visual appearances of the original 3D geometric shape when observed from the viewpoint $\mathbf{c}_k$.

After backbone feature extraction, we feed $\{\mathcal{V}_k\}_{k=1}^{K}$ into a global average pooling (GAP) layer to deduce a collection of vectorized multi-view visual descriptors $\{\mathbf{v}_k \in \mathbb{R}^{C_t}\}_{k=1}^{K}$ that serve as the teacher knowledge to be transferred into the target student model in the actual distillation process.

\subsection{Student Network for 3D Point Cloud Modeling} \label{sec:III.C}

In contrast to the maturity of 2D image modeling networks, 3D deep learning on raw point clouds is still a newly-emerging yet fast-growing research area. Comparatively, existing point cloud learning networks are faced with the following aspects of restrictions:

\begin{itemize}
	\item Due to the difficulty of collecting and labeling 3D shape models, the current geometry community still lacks large-scale richly-annotated 3D shape repositories comparable to their 2D counterparts (e.g., \cite{krizhevsky2012imagenet,lin2014microsoft}).
	\item Limited by the insufficiency of training data, mainstream point cloud networks are actually designed to be far from ``deep'', in order to relieve parameter overfitting.
	\item Different from the 2D vision community where ``pretrain-and-finetune'' has become a default practice, mainstream point cloud networks are still trained from scratch.
\end{itemize}

Consequently, point-based learning paradigms demonstrate insufficient feature learning capacity, and unstable and sub-optimal downstream task performance.

Though there is no unified point feature extraction schemes, we can formulate a general abstraction of the actual behaviors of various existing deep set architectures. Formally, the student deep 3D point encoder $\mathcal{M}_s$ consumes an input set of 3D points and outputs high-dimensional point embedding vectors (from the last abstraction level), which can be described as
\begin{equation} \label{general-student-formulation-1}
	\mathcal{G} = \mathcal{M}_s(\mathcal{P}),
\end{equation}
\noindent where $\mathcal{G} \in \mathbb{R}^{N_a \times C_s}$ represents a set of $C_s$-dimensional feature vectors extracted from the original $N$ input points. The actual learning process operating on $\mathcal{P}$ can be further detailed as
\begin{equation} \label{general-student-formulation-2}
	\mathcal{M}_s(\mathcal{P}) = \varphi_\mathrm{bm}(\mathcal{A}),
\end{equation}
\noindent where $\mathcal{A} \in \mathbb{R}^{N_a \times 3}$ denotes $N_a$ (what we call) feature anchors, and $ \varphi_\mathrm{bm}(\cdot)$ represents a learnable feature mapping that point-wisely embeds each of the 3D feature anchor point (in $\mathcal{A}$) to its corresponding feature vector (in $\mathcal{G}$). Here, we propose the key concept of ``feature anchors'' for convenience of forming a concise and general description for the student behaviors. In practice, depending on specific network architectures and task properties, there exist different definitions and implementations of the feature anchors $\mathcal{A}$. For example, in hierarchical learning frameworks where $N_a < N$, feature anchors $\mathcal{A}$ can be a downsampled subset of input points (as conducted in \cite{qi2017pointnet++,liu2019relation,zhang2019shellnet}) or a sparse set of adaptively learned and generated anchor positions (as conducted in \cite{li2018so,yan2020pointasnl,nezhadarya2020adaptive}). By contrast, many other approaches \cite{qi2017pointnet,hua2018pointwise,wang2019dynamic} also choose to deduce point-wise embeddings for all input points, such that $N_a = N$, which means that the feature anchors $\mathcal{A}$ are exactly the input point cloud $\mathcal{P}$ itself.

\begin{figure*}[t]
	\centering
	\includegraphics[width=0.90\linewidth]{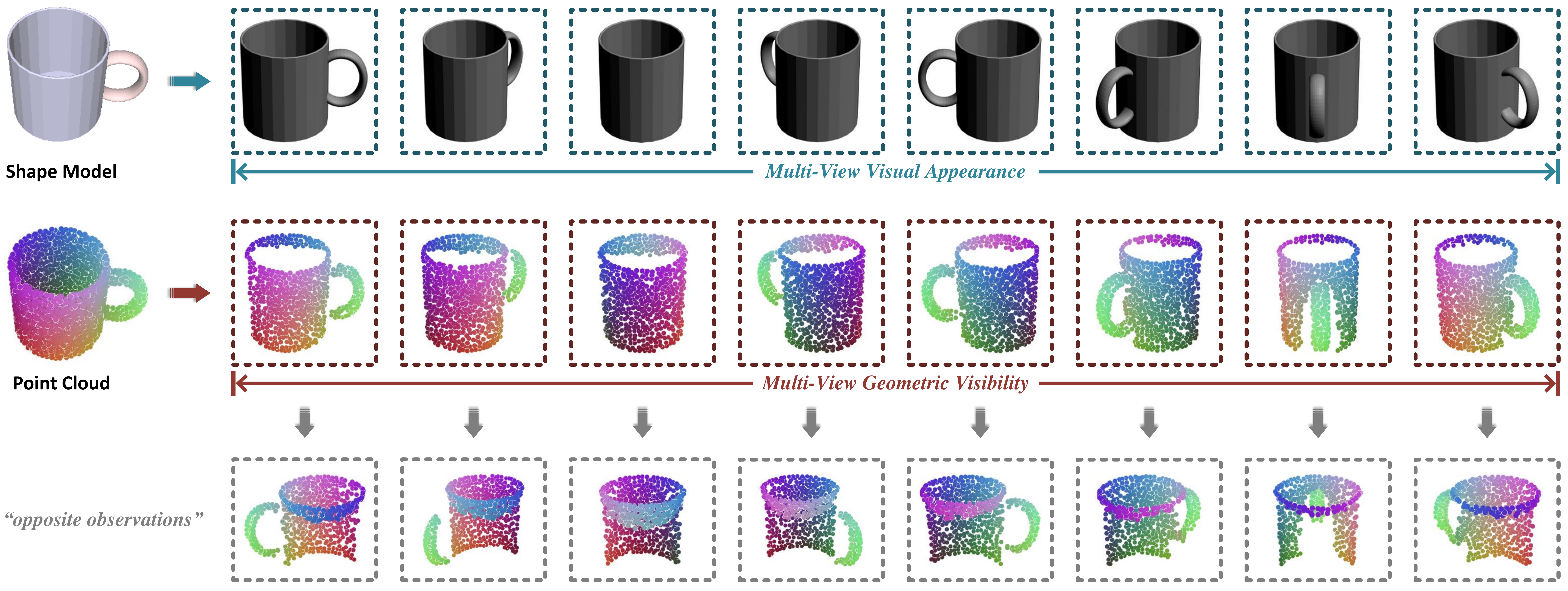}
	\caption{Illustration of multi-view visibility checking, where we apply hidden point removal \cite{katz2007direct} to select visible points observed from the specified viewpoint. In the first and second rows, we respectively display the rendered multi-view images and the selected visible points (subsets of the complete point cloud) generated from the same pre-defined views. For better visual effect, we further provide visualizations of the selected visible points when observed from the opposite azimuth angles in the third row.}
	\label{fig:multi-view-visible-points}
\end{figure*}

\subsection{Multi-View Visibility-aware Feature Alignment} \label{sec:III.D}

Conventional development protocols of knowledge distillation studies mainly focus on homogeneous model architectures (e.g., deep and shallow CNNs) and the same data structure but different contents (e.g., RGB and depth images). Differently, here we are faced with significant heterogeneity of both model and data across 2D and 3D domains. Essentially, our goal is to establish connections in a reasonable manner between the $K$ view-specific visual descriptors $\{\mathbf{v}_k\}_{k=1}^{K}$ (extracted from $\mathcal{M}_t$) corresponding to viewpoints $\{\mathbf{c}_k\}_{k=1}^{K}$ and the $N_a$ anchor embedding vectors in $\mathcal{G}$ (extracted from $\mathcal{M}_s$).

Intuitively, each individual $\mathbf{v}_k$ encodes partial shape appearance that is visible from the viewpoint $\mathbf{c}_k$, while point features of $\mathcal{G}$ capture complete 3D geometry information of the whole shape scope. Such asymmetry strongly drives us to investigate a symmetric visual-geometric feature alignment procedure via view-specific feature projection. Formally, we attempt to distill the teacher knowledge into the student model by imposing the following pair-wise alignment constraint
\begin{equation} \label{distillation-loss}
	\mathcal{L}_\mathrm{dist} = \sum\nolimits_{k=1}^{K} \left\| \mathbf{v}_k - \phi_k(\mathcal{G}) \right\|_1,
\end{equation}
\noindent where $\phi_k(\cdot)$ denotes a learnable feature projector aggregating point embeddings in $\mathcal{G}$ into a vectorized geometric descriptor $\mathbf{g}_k \in \mathbb{R}^{C_t}$, which is supposed to particularly encode the partial shape geometry within the field of view when observed from the corresponding viewpoint $\mathbf{c}_k$. 

Thus, by view-wisely aligning each pair of visual-geometric descriptors in $\{\mathbf{v}_k\}_{k=1}^{K}$ and $\{\mathbf{g}_k = \phi_k(\mathcal{G})\}_{k=1}^{K}$, the multi-view teacher knowledge can be adequately exploited for distillation in an interpretable manner. \\

\noindent \textbf{VAFP Mechanism}. Obviously, under our proposed knowledge transfer framework, the core problem is to design an effective view-specific geometric feature projector $\phi_k$. To this end, here we design a novel VAFP mechanism to generate $\{\mathbf{g}_k\}_{k=1}^{K}$ from point embeddings in $\mathcal{G}$ on the basis of point-wise visibility of feature anchors $\mathcal{A}$.

To implement visibility checking for each of the $N_a$ feature anchor points in $\mathcal{A}$, we employ a simple and fast hidden point removal algorithm \cite{katz2007direct}, i.e., a classic computational approach proposed earlier in the computer graphics community. Fig. \ref{fig:multi-view-visible-points} visually illustrates the selected visible points corresponding to varying manually specified viewpoints. This geometrically-meaningful selection mechanism enables us to aggregate the $N_a$ feature anchor points in $\mathcal{A}$ view-wisely into $K$ independent visible subsets $\{\tilde{\mathcal{A}}_k \in \mathbb{R}^{N_k \times 3}\}_{k=1}^{K}$ with respect to viewpoints $\{\mathbf{c}_k\}_{k=1}^{K}$, where we have $\tilde{\mathcal{A}}_k \subseteq \mathcal{A}$.

Furthermore, based on the one-to-one mapping relationships (i.e., row-wise correspondence) between feature anchor points in $\mathcal{A}$ and their high-dimensional embeddings in $\mathcal{G}$, we further obtain $K$ view-specific clusters of feature embeddings denoted as $\{ \tilde{\mathcal{G}}_k \in \mathbb{R}^{N_k \times C_s} \}$, where we have $\tilde{\mathcal{G}}_k \subseteq \mathcal{G}$. More concretely, for viewpoint $\mathbf{c}_k$, suppose that the $i^\mathrm{th}$ point in $\tilde{\mathcal{A}}_k$ corresponds to the $j^\mathrm{th}$ point in $\mathcal{A}$, then the $j^\mathrm{th}$ embedding vector in $\mathcal{G}$ will be collected into $\tilde{\mathcal{G}}_k$.

Intuitively, the above view-specific feature grouping process simply finds the embedding vectors of the visible anchor points and then collect them together. After that, we can deduce the desired multi-view geometric descriptors $\{\mathbf{g}_k\}_{k=1}^{K}$ by
\begin{equation} \label{mv-geo-descriptors}
	\mathbf{g}_k = \phi_k(\mathcal{G}) = \hbar_\mathrm{align}(\rho_\mathrm{max}(\tilde{\mathcal{G}}_k)),
\end{equation}
\noindent where $\rho_\mathrm{max}$ means applying channel-wise max-pooling on $\tilde{\mathcal{G}}_k$ to generate a vectorized $C_s$-dimensional codeword, and $\hbar_\mathrm{align}$ is implemented as a single fully-connected (FC) layer, which aligns the number of feature channels (from $C_s$ to $C_t$) so that we can impose $L_1$ loss between $\mathbf{v}_k$ and $\mathbf{g}_k$ (as formulated in Eq. (\ref{distillation-loss})). From a different perspective, the existence of $\hbar_\mathrm{align}$ also indicates that we tend to perform visual-geometric feature alignment in another hidden space, instead of directly forcing point cloud features to be close to image features.

\subsection{Training Strategy} \label{sec:III.E}

Just like standard knowledge distillation pipelines, given a specific 3D shape recognition task, we start by pretraining the teacher branch $\mathcal{B}_t$ on 2D multi-view images before distillation. After that, we train the target student branch $\mathcal{B}_s$ on 3D point clouds while in the meantime transferring knowledge exported from the fixed deep image encoder $\mathcal{M}_t$ into the deep point cloud encoder $\mathcal{M}_s$.

The overall training objective $\mathcal{L}_\mathrm{overall}$ can be formulated as a weighted summation of the task loss $\mathcal{L}_\mathrm{task}$ and the distillation constraint $\mathcal{L}_\mathrm{dist}$ (Eq. (\ref{distillation-loss})) as given in the following
\begin{equation} \label{overall-training-objective}
	\mathcal{L}_\mathrm{overall} = \omega_t \cdot \mathcal{L}_\mathrm{task} + \omega_d \cdot \mathcal{L}_\mathrm{dist},
\end{equation}
\noindent where we empirically set $\omega_t = 0.1$ and $\omega_d = 1 / K$ in all our experiments. Functionally, $\mathcal{L}_\mathrm{task}$ drives the optimization of the whole student learning branch, including both the encoder $\mathcal{M}_s$ and the head $\mathcal{H}_s$, while $\mathcal{L}_\mathrm{dist}$ only has impact on the encoder $\mathcal{M}_s$ by constraining its feature outputs. In particular, for some deep point cloud encoders (e.g., \cite{qi2017pointnet}) that require optimizing auxiliary regularizers, we uniformly conform to their original weighting factors applied on regularization terms.

\begin{figure*}[t]
	\centering
	\includegraphics[width=0.90\linewidth]{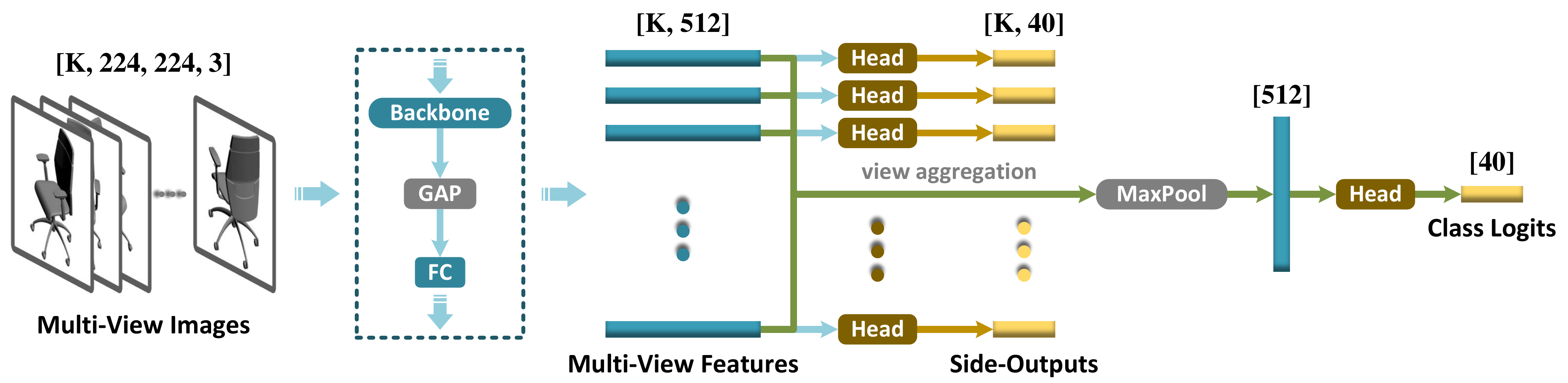}
	\caption{Illustration of the teacher network pretrained for 2D multi-view image-based 3D shape classification. Given a collection of multi-view images at the input end, we extract the corresponding vectorized ($512$-dimensional) visual descriptors, which are aggregated via view pooling \cite{su2015multi} and then mapped to predict category scores. In addition, we make independent predictions from each single-view descriptor via side-output supervision.}
	\label{fig:teacher-branch-cls}
\end{figure*}

\section{Experiments} \label{sec:experiments}

We conducted experiments to validate the effectiveness and universality of the proposed PointMCD architecture on three 3D shape recognition scenarios, including shape classification (Sec. \ref{sec:exp-cls}), part segmentation (Sec. \ref{sec:exp-seg}), and unsupervised learning (Sec. \ref{sec:exp-uns}). Within each sub-section, we introduced the corresponding benchmark dataset, evaluation protocol, and detailed technical implementation. In Sec. \ref{sec:exp-abl}, we provided more comprehensive comparisons and ablation experiments.

\begin{table}[t]
	\centering	
	\renewcommand\arraystretch{1.30}
	\setlength{\tabcolsep}{13.5pt}
	\caption{Performance comparison of shape classification on ModelNet40 dataset in terms of overall accuracy without any inference voting. Note that ``$\ast$'' means normal vectors are exploited as additional inputs.}
	\begin{tabular}{ l l l }
		\toprule[1.2pt]
		\textbf{Method} & \textbf{Data Modality} & \textbf{OAcc (\%)} \\
		\hline
		VoxNet \cite{qi2017pointnet} & Voxels; 32$^3$ & 85.9 \\
		O-CNN \cite{wang2017cnn} & Voxels; 64$^3$ & 89.9 \\
		VRN-Single \cite{brock2016generative} & Voxels; 32$^3$ & 91.3 \\
		\hline
		MVCNN \cite{su2015multi} & Images; 80$\times$ & 90.1 \\
		MVCNN-New \cite{su2018deeper} & Images; 12$\times$ & 95.0 \\
		RotationNet \cite{kanezaki2018rotationnet} & Images; 20$\times$ & 97.4 \\
		View-GCN \cite{wei2020view} & Images; 20$\times$ & 97.6 \\
		\hline
		PointNet \cite{qi2017pointnet} & Points; 1K & 89.2 \\
		PointNet++ \cite{qi2017pointnet++} & Points; 1K & 90.7 \\
		PointNet++ \cite{qi2017pointnet++} $\mathbf{\ast}$ & Points; 5K & 91.9 \\
		SpiderCNN \cite{xu2018spidercnn} $\mathbf{\ast}$ & Points; 1K & 92.4 \\
		SO-Net \cite{li2018so} $\mathbf{\ast}$ & Points; 5K & 93.4 \\
		KPConv \cite{thomas2019kpconv} & Points; 7K & 92.9 \\
		DGCNN \cite{wang2019dynamic} & Points; 1K & 92.9 \\
		RS-CNN \cite{liu2019relation} & Points; 1K & 92.9 \\
		PAConv \cite{xu2021paconv} & Points; 1K & 93.6 \\
		{[ST]} Point-BERT \cite{yu2022point} & Points; 8K & 93.8 \\
		CurveNet \cite{xiang2021walk} & Points; 1K & 93.8 \\
		{[ST]} Point-MAE \cite{pang2022masked} & Points; 8K & 94.0 \\
		PointMLP \cite{ma2022rethinking} & Points; 1K & 94.1 \\
		\hline\hline
		\textbf{\textit{MCD-PointNet}} & Points; 1K & \textbf{\textit{91.1 (+1.9)}} \\
		\textbf{\textit{MCD-PointNet++}} & Points; 1K & \textbf{\textit{93.3 (+2.6)}} \\
		\textbf{\textit{MCD-DGCNN}} & Points; 1K & \textbf{\textit{93.7 (+0.8)}} \\
		\textbf{\textit{MCD-CurveNet}} & Points; 1K & \textbf{\textit{94.3 (+0.5)}} \\
		\bottomrule[1.2pt]
	\end{tabular}
	\label{tab:modelnet40-classification}
\end{table}

\subsection{Evaluations on Shape Classification} \label{sec:exp-cls}

As one of the most fundamental benchmark tasks, 3D shape classification has been widely adopted to evaluate the learning capacity of deep set architectures. Generally, this task aims at extracting from an input object point cloud a vectorized global shape signature, which can be further mapped through a stack of FC layers to predict category scores.

We conducted experiments on the ModelNet40 \cite{wu20153d} dataset, which totally consists of $12311$ synthetic mesh models of man-made 3D objects covering $40$ semantic categories. Following the official split, we have $9843$ shapes for training and the rest $2468$ for testing. We applied Poisson disk sampling (PDS) to uniformly discretize $1024$ spatial points from the original mesh faces. Note that we did not preserve ground-truth point-wise normals as additional input attributes.

For the teacher learning branch, we adopted a mature multi-view learning framework (i.e., MVCNN \cite{su2015multi}) built upon multi-input CNNs while making a few modifications to both encoder $\mathcal{M}_t$ and head $\mathcal{H}_t$ for performance boost. Fig. \ref{fig:teacher-branch-cls} illustrates the overall workflow and core components of the teacher learning branch $\mathcal{B}_t$ for multi-view image-based 3D shape classification. Specifically, we introduced MobileNetV2 \cite{sandler2018mobilenetv2}, a light-weight 2D CNN backbone network, as the teacher image encoder $\mathcal{M}_t$. After backbone feature extraction, view-wise visual descriptors are on one hand aggregated to predict the final class logits and on the other hand separately mapped to form side-outputs. For the rendering pipeline, we used the same processing procedure as recommenced in \cite{su2018deeper}, in which $K = 12$, $\theta^{az}_k \in \{ k \pi / 6 \}_{k=1}^{12}$, $\theta^{el}_k = \pi / 6$, and $\mu = 1$. After pretraining, the resulting teacher branch achieves $96.7\%$ overall classification accuracy on the ModelNet40 testing set, and its backbone encoder parameters are fixed when applied to distill different students.

In practice, we selected four representative deep point cloud encoders as the target 3D student models (i.e., $\mathcal{M}_s$), including three classic learning frameworks (PointNet \cite{qi2017pointnet}, PointNet++ \cite{qi2017pointnet++}, DGCNN \cite{wang2019dynamic}) and a more recent state-of-the-art method CurveNet \cite{xiang2021walk}. It is worth mentioning that, during performance comparison, we did not involve the results \cite{liu2019relation,xu2021paconv,xiang2021walk,ma2022rethinking} obtained by extensively voting a large amount of trials during inference, which turns to be unstable and highly cumbersome.

\begin{figure*}[t]
	\centering
	\includegraphics[width=0.90\linewidth]{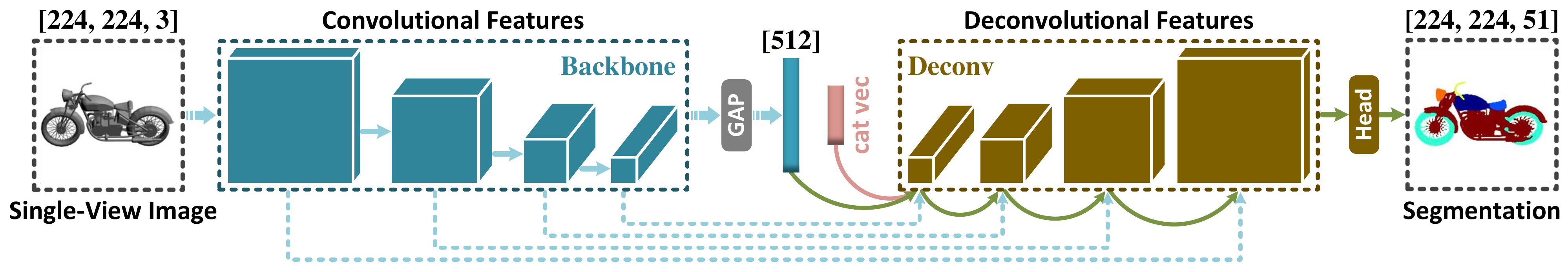}
	\caption{Illustration of the teacher network pretrained for semantic segmentation of 2D rendered images. Instead of jointly learning from multiple views, here we independently perform segmentation on each single view image. After extracting a vectorized ($512$-dimensional) visual descriptor from the input single-view image, we append an auxiliary categorical vector to indicate object-level semantics and further deploy a stack of deconvolutional layers, following the classic U-Net-style \cite{ronneberger2015u} encoder-decoder network design, to predict pixel-wise labels on a full-resolution segmentation map.}
	\label{fig:teacher-branch-seg}
\end{figure*}

Table \ref{tab:modelnet40-classification} quantitatively compares classification performances of different modeling paradigms and specific learning models, from which we can draw several aspects of useful conclusions. First, it is observed that, despite the continuous progress made by the point cloud community, image-based visual modeling still outperforms voxel-based and point-based paradigms with significant margins. Second, it turns out that PointMCD is able to produce stable and obvious performance boost for various types of deep point encoders. Notably, the distilled PointNet++ and DGCNN are boosted to achieve $93.3\%$ and $93.7\%$ overall accuracy, which are comparable to more complicated network structures investigated in follow-up works. Even for CurveNet, a much more powerful baseline model that originally achieves $93.8\%$ state-of-the-art performance, our distillation framework still brings satisfactory gains. Third, an interesting observation is that the actual performance gap between some earlier classic methods (e.g., \cite{qi2017pointnet++,wang2019dynamic}) and more recently proposed methods (e.g., \cite{xiang2021walk,ma2022rethinking}) has been largely reduced via the distillation of PointMCD. This phenomenon implies that the actual potential of some previous network design paradigms for point cloud learning may not have been sufficiently explored and realized.

\begin{table}[t]
	\centering	
	\renewcommand\arraystretch{1.30}
	\setlength{\tabcolsep}{22.5pt}
	\caption{Performance comparison of object part segmentation on ShapeNetPart dataset under the measurement of instance-averaged mean intersection-over-union. Note that ``$\ast$'' means normal vectors are exploited as additional inputs.}
	\begin{tabular}{ l l }
		\toprule[1.2pt]
		\textbf{Method} & \textbf{mIoU (\%)} \\
		\hline
		PointNet \cite{qi2017pointnet} & 83.7 \\
		PointNet++ \cite{qi2017pointnet++} $\mathbf{\ast}$ & 85.1 \\
		SpiderCNN \cite{xu2018spidercnn} $\mathbf{\ast}$ & 85.3 \\
		SO-Net \cite{li2018so} $\mathbf{\ast}$ & 84.9 \\
		KPConv \cite{thomas2019kpconv} & 86.4 \\
		DGCNN \cite{wang2019dynamic} & 85.1 \\
		RS-CNN \cite{liu2019relation} & 85.8 \\
		PAConv \cite{xu2021paconv} & 86.0 \\
		CurveNet \cite{xiang2021walk} & 86.6 \\
		PointMLP \cite{ma2022rethinking} & 86.1 \\
		\hline\hline
		\textbf{\textit{MCD-PointNet}} & \textbf{\textit{85.9 (+2.2)}} \\
		\textbf{\textit{MCD-PointNet++}} & \textbf{\textit{86.4 (+1.3)}} \\
		\textbf{\textit{MCD-DGCNN}} & \textbf{\textit{86.8 (+1.7)}} \\
		\textbf{\textit{MCD-CurveNet}} & \textbf{\textit{87.1 (+0.5)}} \\
		\bottomrule[1.2pt]
	\end{tabular}
	\label{tab:shapenetpart-segmentation}
\end{table}

\subsection{Evaluations on Part Segmentation} \label{sec:exp-seg}

\begin{figure*}[t]
	\centering
	\includegraphics[width=0.90\linewidth]{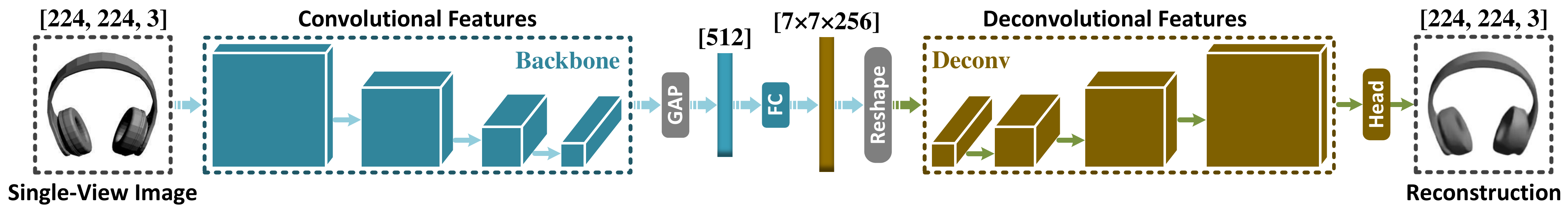}
	\caption{Illustration of the teacher network pretrained for single-view image reconstruction. The overall learning framework is designed as a standard deep convolutional image auto-encoder. The encoding stage represents an input image into a compact $512$-dimensional feature vector. After feature channel lifting and reshaping, the decoding phase further reconstructs a full-resolution image.}
	\label{fig:teacher-branch-uns}
\end{figure*}

To further demonstrate the effectiveness of our approach on more fine-grained per-point semantic understanding scenarios, we experimented with part segmentation of 3D objects on the ShapeNetPart \cite{yi2016scalable} dataset, consisting of $16881$ point-wisely annotated 3D object models covering totally $50$ different parts defined on $16$ object categories. Following common evaluation protocols, there exist $14007$ training models and $2874$ testing models, each of which contains $2048$ labeled spatial points. Similar to our preceding shape classification experiments, here we did not use ground-truth point-wise normals as additional input attributes, and our inference performance was obtained without any voting strategy.

For the teacher learning branch, it is worth mentioning that, in contrast to global geometry understanding applications such as classification/retrieval in which multi-view 2D image-based processing architectures have already been richly investigated, there is relatively little such effort on 3D shape segmentation scenarios. Hence, instead of resorting to existing frameworks, here we simply designed a standard 2D encoder-decoder \cite{ronneberger2015u} convolutional architecture that consumes an input single-view image and accordingly predict a full-resolution segmentation map, as illustrated in Figure \ref{fig:teacher-branch-seg}. More specifically, we selected VGG11 \cite{simonyan2014very} as backbone image encoder $\mathcal{M}_t$ while removing the last spatial max-pooling layer to enlarge the final feature map resolution. Following common practice in previous works on part segmentation, we inserted a categorical vector between the encoded vectorized visual descriptors and the subsequent deconvolutional layers to supplement object-level semantics. The rendering pipeline basically inherits our preceding shape classification experiments, while we additionally expanded the interval of azimuth angles from $\pi/6$ to $\pi/4$ and further added another elevation angle of $-\pi/6$, such that $K=16$, $\theta^{az}_k \in \{ k \pi / 4 \}_{k=1}^{8}$, $\theta^{el}_k = \{\pm pi/6\}$, and $\mu = 1$. For the generation of pixel-wisely annotated ground-truth maps, we projected point-wise labels onto 2D image planes, where another new category of ``background'' that corresponds to empty pixels during the rendering process need to be added into the original annotation set. During our explorations, we did not quantitatively evaluate the actual segmentation performance of such a teacher learning branch, because there is no viable way to determine semantic labels of interior points located inside shape surface.

We experimented with the same four representative deep set architectures \cite{qi2017pointnet,qi2017pointnet++,wang2019dynamic,xiang2021walk} as selected in our preceding shape classification experiments. As reported in Table \ref{tab:shapenetpart-segmentation}, our PointMCD consistently brings obvious performance improvement. In particular, PointNet can be significantly boosted from the original $83.7\%$ to the distilled $85.9\%$, which is on par with more advanced frameworks \cite{liu2019relation,xu2021paconv,ma2022rethinking}. After distillation, the classic DGCNN framework also reaches $86.8\%$ with large gains of $1.7\%$, which even outperforms the original CurveNet. For the most powerful CurveNet baseline, our approach still brings $0.5\%$ performance boost. These results demonstrate the effectiveness of our method on fine-grained per-point labeling tasks, despite that the actual distillation process only operates on global feature vectors without pixel-to-point alignment.

\subsection{Evaluations on Unsupervised Learning} \label{sec:exp-uns}

\begin{table}[t]
	\centering	
	\renewcommand\arraystretch{1.30}
	\setlength{\tabcolsep}{10.0pt}
	\caption{Performance comparison of unsupervised learning (transfer classification) on ModelNet40 dataset. Note that ``$\mathbf{\dag}$'' means distilling visual knowledge from an ImageNet pretrained backbone image encoder without task-specific pretraining of the whole teacher learning branch on the source dataset. Particularly, multiple different point cloud backbones are involved in \cite{huang2021spatio,afham2022crosspoint,chen2021shape,eckart2021self}, and below we report the results obtained from the PointNet \cite{qi2017pointnet} backbone, whose representation power is basically close to the point cloud encoder used in FoldingNet \cite{yang2018foldingnet}.}
	\begin{tabular}{ l l l }
		\toprule[1.2pt]
		\textbf{Method} & \textbf{Data Modality} & \textbf{OAcc (\%)} \\
		\hline
		TL-Net \cite{girdhar2016learning} & Voxels; 20$^3$ & 74.4 \\
		VConv-DAE \cite{sharma2016vconv} & Voxels; 30$^3$ & 75.5 \\
		3D-GAN \cite{wu2016learning} & Voxels; 64$^3$ & 83.3 \\
		\hline
		VIP-GAN \cite{han2019view} & Images; 12$\times$ & 90.2 \\
		\hline
		Latent-GAN \cite{achlioptas2018learning} & Points; 2K & 85.7 \\
		STRL \cite{huang2021spatio} & Points; 2K & 88.3 \\
		FoldingNet \cite{yang2018foldingnet} & Points; 2K & 88.4 \\
		CrossPoint \cite{afham2022crosspoint} & Points; 2K & 89.1 \\
		3D-PointCapsNet \cite{zhao20193d} & Points; 2K & 89.3 \\
		SelfContrast \cite{du2021self} & Points; 2K & 89.6 \\
		GTIF \cite{chen2019deep} & Points; 2K & 89.7 \\
		SelfCorrection \cite{chen2021shape}  & Points; 2K & 89.9 \\
		ParAE \cite{eckart2021self} & Points; 1K & 90.3 \\
		GSIR \cite{chen2021unsupervised} & Points; 1K & 90.4 \\
		\hline\hline
		\textbf{\textit{MCD-FoldingNet}} & Points; 2K & \textbf{\textit{90.0 (+1.6)}} \\
		\textbf{\textit{MCD-FoldingNet}} $\mathbf{\dag}$ & Points; 2K & \textbf{\textit{89.8 (+1.4)}} \\
		\bottomrule[1.2pt]
	\end{tabular}
	\label{tab:modelnet40-unsupervised-learning}
\end{table}

Previous experiments on classification and segmentation are targeted at supervised learning scenarios with requirements of domain-specific data annotations. Taking a step forward, here we further explored the possibility of transferring generic 2D visual knowledge extracted from unsupervised image feature learning frameworks to boost 3D geometric modeling. Practically, we adopted one of the most popular evaluation protocols (known as transfer classification) employed by many previous works \cite{yang2018foldingnet,chen2019deep,zhao20193d}. The overall evaluation protocol relies on a source dataset (i.e., ShapeNetCore \cite{chang2015shapenet}) for representation learning in an unsupervised manner and another target dataset (i.e., ModelNet40 \cite{wu20153d}) for linear probing. More specifically, the first step is to pretrain the unsupervised learning model on the whole source dataset. Then, we directly employ the trained model to convert each of the shape models in the whole target dataset into a global feature vector, during which the network parameters are fixed. After that, we train and test a linear support vector machine (SVM) classifier using the extracted feature vectors of all training and testing models. In our experiments, each point cloud in both the source and the target datasets contains $2048$ spatial points uniformly discretized from the original mesh faces.

We selected FoldingNet \cite{yang2018foldingnet}, a well-known reconstruction-driven unsupervised point cloud representation learning model, as the student branch, which consists of a graph-based encoder (i.e., $\mathcal{M}_s$) and a folding-based decoder (i.e., $\mathcal{H}_s$). Accordingly, we designed a standard convolutional auto-encoder for single image reconstruction, as illustrated in Figure \ref{fig:teacher-branch-uns}. In the encoding stage, we deployed the same backbone encoder as adopted in our preceding part segmentation experiments. The decoding stage comprises a stack of deconvolutional layers to generate a full-resolution image. The rendering pipeline is the same as described in Sec. \ref{sec:exp-seg} to generate multi-view images on the source dataset of ShapeNetCore.

\begin{figure}[t]
	\centering
	\includegraphics[width=0.995\linewidth]{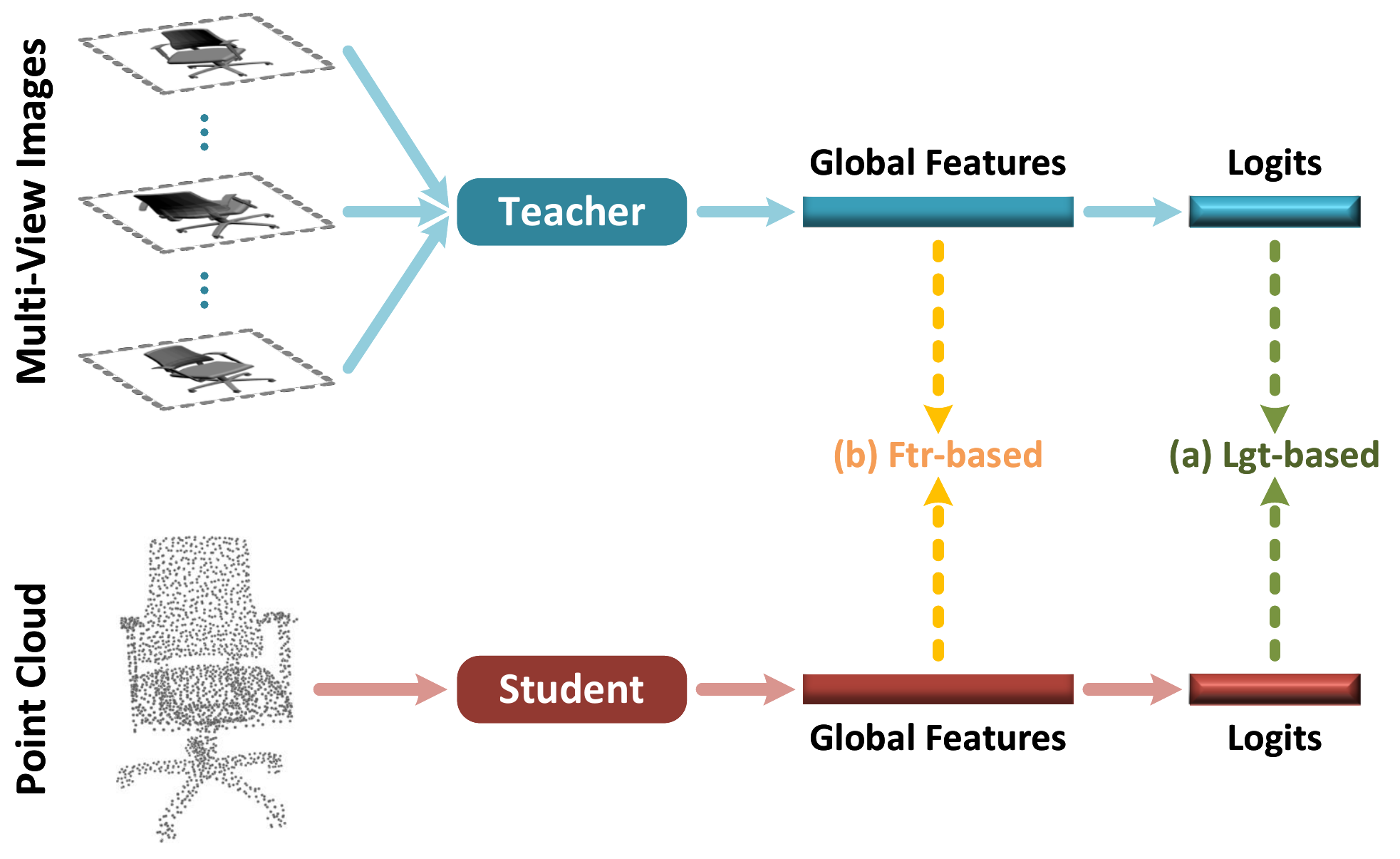}
	\caption{Illustration of two conventional (a) logit-based and (b) feature-based distillation baselines, which impose alignment constraints over the final class logits and the intermediate global feature vectors, respectively.}
	\label{fig:conventional-distillation}
\end{figure}

Table \ref{tab:modelnet40-unsupervised-learning} quantitatively compares transfer classification performances of different learning frameworks. It is observed that multi-view image-based modeling paradigm can still achieve dominating overall accuracy. As a representative point-based unsupervised learning model, FoldingNet also performs favorably against other more advanced state-of-the-art approaches. After distillation, its accuracy is boosted from $88.4\%$ to $90.0\%$, which is comparable to \cite{han2019view} and outperforms \cite{zhao20193d,chen2019deep}.

Additionally, we are interested in the potential of distilling generic visual knowledge learned from natural image statistics without task-specific pretraining for the teacher model. To this end, we did not pretrain the whole teacher learning branch on the source dataset for single image reconstruction. Instead, we employed the original VGG11 backbone image encoder with ImageNet pretrained model parameters to produce the teacher knowledge during the distillation process. As shown in the last row of Table \ref{tab:modelnet40-unsupervised-learning}, this variant suffers from slight performance degradation from $90.0\%$ to $89.8\%$, but still turns to be highly competitive and encouraging, considering the big domain gap between natural images and shape renderings.

\subsection{Ablation Study} \label{sec:exp-abl}

\begin{table}[t]
	\centering	
	\renewcommand\arraystretch{1.30}
	\setlength{\tabcolsep}{8.0pt}
	\caption{Performance verification of conventional logit-based and feature-based distillation baselines.}
	\begin{tabular}{ l c | c c}
		\toprule[1.2pt]
		\textbf{Method} & \textbf{PointMCD} & \textbf{Lgt-based} & \textbf{Ftr-based} \\
		\hline
		\textbf{\textit{MCD-PointNet}} & $91.1$ & $90.2$ & $90.4$ \\
		\textbf{\textit{MCD-PointNet++}} & $93.3$ & $92.5$ & $92.8$ \\
		\textbf{\textit{MCD-DGCNN}} & $93.7$ & $93.0$ & $93.2$ \\
		\textbf{\textit{MCD-CurveNet}} & $94.3$ & $93.8$ & $93.9$ \\
		\bottomrule[1.2pt]
	\end{tabular}
	\label{tab:abs-distillation-baselines}
\end{table}

In order to facilitate more comprehensive understanding of the whole processing pipeline of the proposed PointMCD, we conducted ablation studies on ModelNet40 classification for distilling the preceding four student architectures \cite{qi2017pointnet,qi2017pointnet++,wang2019dynamic,xiang2021walk} under different experimental conditions. 

First, we constructed and evaluated two conventional distillation paradigms to validate the necessity and superiority of our PointMCD that is particularly customized for multi-view image-to-point distillation. Second, we investigated different hyperparameter configurations and their influences to the final performance of the whole processing pipeline. Third, one more important experiment was added to evaluate the performance of purely point-deduced baseline pipelines, which can strongly reveal the great potential of transferring knowledge from 2D image domains to 3D point cloud domains.

\subsubsection{Conventional distillation paradigm}

In the research area of knowledge distillation, there are two representative learning paradigms, namely response-based \cite{hinton2015distilling} and feature-based \cite{romero2014fitnets} methods. The former type of works aims at performing vector alignment between the final class logits outputted from the last layers of teacher and student learning branches. For the latter type of works, attentions are shifted from output distribution mimicking to feature-level alignment within intermediate network layers. 

Here, as illustrated in Figure \ref{fig:conventional-distillation}, we accordingly adapted the above two classic distillation paradigms into our development protocol, leading to the logit-based and feature-based baseline frameworks. Table \ref{tab:abs-distillation-baselines} compares performance of the proposed PointMCD against these two baselines. First, our experimental results strongly validate that multi-view image-to-point knowledge transfer serves as a robust and stable way of empowering various point cloud learning frameworks. Note that even the most straightforward distillation approach can bring different degrees of task performance enhancement in almost all of our experimental setups. Second, it turns out that the feature-level teacher guidance is more informative than the soft targets (i.e., logits), based on the observation that the feature-based variants uniformly outperform their logit-based counterparts.

\subsubsection{Hyperparameter influence}
There are two critical influencing factors within the overall processing pipeline, i.e., 1) the number of views during the rendering process (i.e., $K$); 2) the weighting scheme between task loss and distillation loss used to formulate the overall training objective (i.e., Eq. (\ref{overall-training-objective})).

\begin{table}[t]
	\centering	
	\renewcommand\arraystretch{1.30}
	\setlength{\tabcolsep}{7.0pt}
	\caption{Ablation studies on the influence of different number of views during multi-view image rendering.}
	\begin{tabular}{ l c | c c c }
		\toprule[1.2pt]
		\textbf{Method} & \textbf{Comp-12} & \textbf{Redu-6} & \textbf{Redu-4} & \textbf{Rand-1} \\
		\hline
		\textbf{\textit{MCD-PointNet}} & $91.1$ & $91.0$ & $90.8$ & $90.7$ \\
		\textbf{\textit{MCD-PointNet++}} & $93.3$ & $93.0$ & $92.9$ & $92.7$ \\
		\textbf{\textit{MCD-DGCNN}} & $93.7$ & $93.4$ & $93.2$ & $93.3$ \\
		\textbf{\textit{MCD-CurveNet}} & $94.3$ & $94.1$ & $94.0$ & $94.0$ \\
		\bottomrule[1.2pt]
	\end{tabular}
	\label{tab:abs-number-of-views}
\end{table}

\begin{table}[t]
	\centering	
	\renewcommand\arraystretch{1.30}
	\setlength{\tabcolsep}{8.0pt}
	\caption{Ablation studies on the influence of different task loss weights within the overall training objective.}
	\begin{tabular}{ l c | c c}
		\toprule[1.2pt]
		\textbf{Method} & $\mathbf{\boldsymbol{\omega}_t=0.1}$ & $\mathbf{\boldsymbol{\omega}_t=0.01}$ & $\mathbf{\boldsymbol{\omega}_t=1.0}$ \\
		\hline
		\textbf{\textit{MCD-PointNet}} & $91.1$ & $81.9$ & $90.8$ \\
		\textbf{\textit{MCD-PointNet++}} & $93.3$ & $83.5$ & $92.9$ \\
		\textbf{\textit{MCD-DGCNN}} & $93.7$ & $87.1$ & $93.2$ \\
		\textbf{\textit{MCD-CurveNet}} & $94.3$ & $89.6$ & $93.9$ \\
		\bottomrule[1.2pt]
	\end{tabular}
	\label{tab:abs-loss-weights}
\end{table}

First, as reported in Table \ref{tab:abs-number-of-views}, we experimented with different number of views. During the generation of multi-view images, we reduced the number of virtual cameras from the complete setting of $K=12$ to $K=\{6, 4, 1\}$, leading to three variants, namely: Redu-6, Redu-4, and Rand-1. More specifically, under the same elevation angle $\theta^{el}_k=\-pi/6$ and observation distance $\mu=1$, we changed azimuth angles to $\theta^{az}_k \in \{ k \pi / 3 \}_{k=1}^{6}$ and $\theta^{az}_k \in \{ k \pi / 2 \}_{k=1}^{4}$ in Redu-6 and Redu-4, respectively. As for Rand-1, throughout the whole training process, we randomly selected one of the original $12$ views at each iteration. We can observe from the quantitative comparisons that using fewer rendering views inevitably causes performance degradation, owing to the loss of visual information captured from those missing viewpoints. Particularly, the Rand-1 variant with the random view selection strategy also shows satisfactory performance, which is viewed as a good trade-off between training cost and inference accuracy.

Second, to investigate the influences of different weighting schemes in the overall training objective $\mathcal{L}_\mathrm{overall}$, we maintained $\omega_d=1/K$ while experimenting with different values of $\omega_t$. As reported in Table \ref{tab:abs-loss-weights}, $\omega_t=0.1$ turns to be a robust choice that uniformly leads to the optimal performance on all the distilled models. Given a much smaller value of $0.01$, the task loss cannot adequately converge, resulting in significant performance degradation. Inversely, enlarging its value to $1.0$ weakens the constraining effect of the distillation loss, which also causes relatively sub-optimal results.

\begin{figure}[t]
	\centering
	\includegraphics[width=0.995\linewidth]{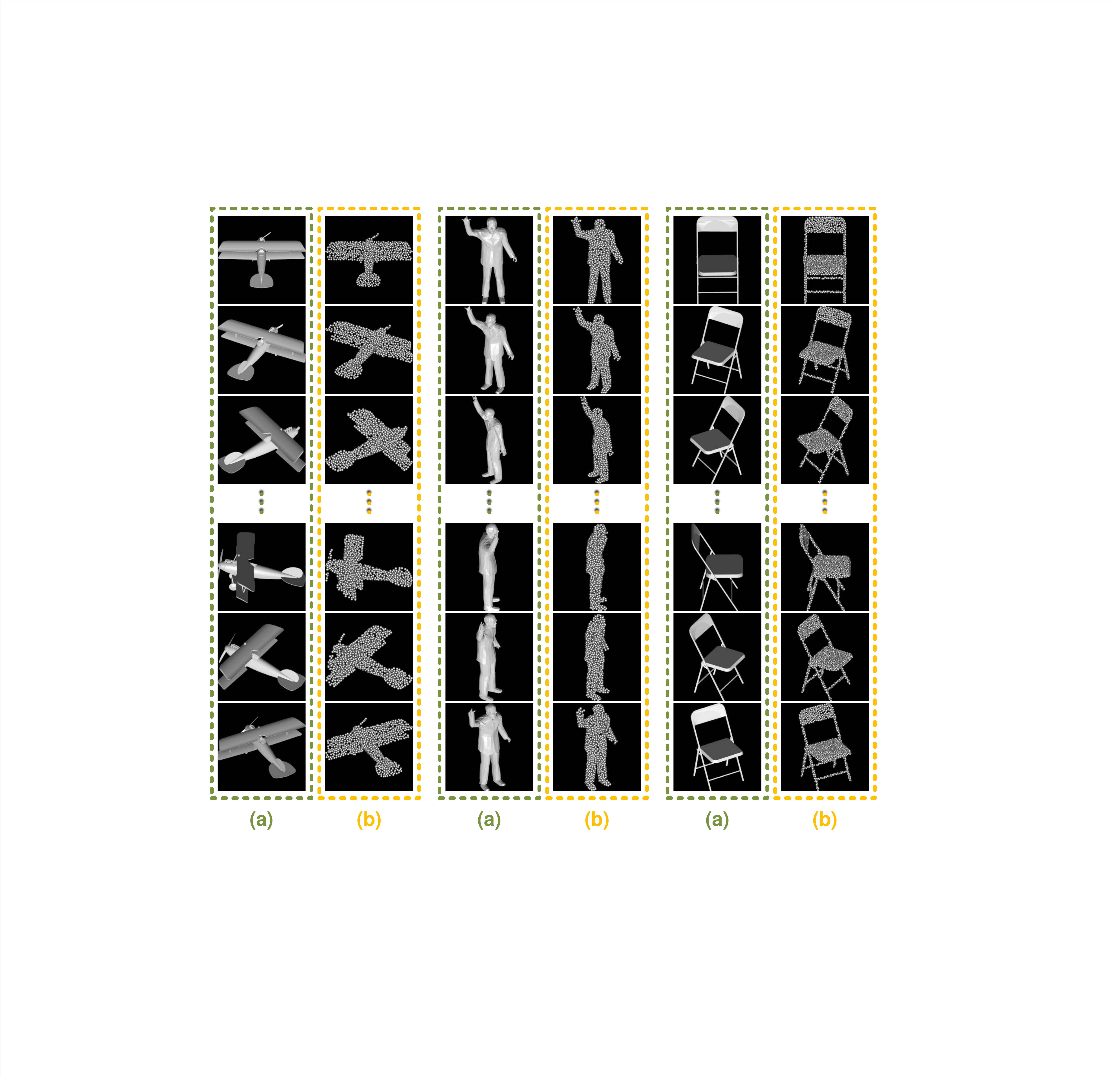}
	\caption{Visual comparison of (a) mesh-based and (b) point-based multi-view rendering pipelines, where the resulting images are generated from polygon meshes and raw point clouds, respectively.}
	\label{fig:rendering-comparison}
\end{figure}

\subsubsection{Purely point-deduced performance}

In all our preceding experiments, there are two aspects of auxiliary data used in the training process. First, off-the-shelf deep image encoders are pretrained on ImageNet, which means implicit exploitation of large-scale labeled image data. Second, multi-view rendered images are computed from manually created polygon meshes, which contain more informative surface properties than point clouds. Although we emphasize that, while finishing training, the actual inference phase is a purely point-based architecture without requirement of any other auxiliary inputs, we are still interested in the possibility of building our processing pipeline completely upon point clouds. To this end, we correspondingly made the following two aspects of modifications to our original settings for constructing point-deduced distillation baselines:

\begin{itemize}
	\item we rendered multi-view images from raw point clouds, rather than polygon meshes. Figure \ref{fig:rendering-comparison} visually compares mesh-based and point-based rendering results; and
	\item we did not initialize the deep image encoder (i.e., $\mathcal{M}_t$) with ImageNet-pretrained network parameters during the pretraining process of the whole teacher learning branch.
\end{itemize}

Table \ref{tab:abs-point-deduced-baselines} quantitatively compares our original framework (i.e., M-Rend) and two point-deduced baselines, in which the first modification produces the P-Rend variant, combining the above two modifications produces the P-Rend-Scratch variant. From these results, we can draw some important conclusions that are worth further emphasizing:

\begin{itemize}
	\item Comparing M-Rend and P-Rend, we can observe that the overall performance degradation is actually insignificant. In particular, for the distillation of PointNet, P-Rend even slightly outperforms the original M-Rend. This indicates that PointMCD can be implemented in a more practical and flexible way without being excessively demanding on the preceding rendering process.
	\item Comparing P-Rend and P-Rend-Scratch, we surprisingly notice that pretraining teachers from scratch is comparable to fine-tuning from ImageNet-pretrained parameters. For the distillation of PointNet++ and CurveNet, P-Rend-Scratch is inferior to P-Rend with performance decrease of $0.1\%$ and $0.2\%$, respectively. In the meanwhile, for the distillation of PointNet and DGCNN, P-Rend-Scratch instead outperforms P-Rend with accuracy boost of $0.3\%$ and $0.1\%$, respectively. We reason that this phenomenon reveals the great potential of learning from heterogeneous feature representations of cross-modal data.
\end{itemize}

Overall, we remind that the above ablative experiments and the corresponding purely point-based performances produced by P-Rend-Scratch are critical and informative, demonstrating that our PointMCD still achieves significant performance boost without any other auxiliary information being introduced in the overall processing pipeline.

\begin{table}[t]
	\centering	
	\renewcommand\arraystretch{1.30}
	\setlength{\tabcolsep}{7.0pt}
	\caption{Performance comparison of our original mesh-deduced PointMCD and its two point-deduced baselines.}
	\begin{tabular}{ l c | c c }
		\toprule[1.2pt]
		\textbf{Method} & \textbf{M-Rend} & \textbf{P-Rend} & \textbf{P-Rend-Scratch} \\
		\hline
		\textbf{\textit{MCD-PointNet}} & $91.1$ & $91.2$ & $91.5$ \\
		\textbf{\textit{MCD-PointNet++}} & $93.3$ & $93.0$ & $92.9$ \\
		\textbf{\textit{MCD-DGCNN}} & $93.7$ & $93.5$ & $93.6$ \\
		\textbf{\textit{MCD-CurveNet}} & $94.3$ & $94.2$ & $94.0$ \\
		\bottomrule[1.2pt]
	\end{tabular}
	\label{tab:abs-point-deduced-baselines}
\end{table}

\subsubsection{Evaluations on real-scanned data} 

The preceding experiments use datasets \cite{wu20153d,yi2016scalable,chang2015shapenet} of synthetic shape models. Here, we further evaluated the effectiveness of our approach on the real-scanned object classification dataset of ScanObjectNN \cite{uy2019revisiting}. Since mesh models are not available in this dataset, we used point-based rendering for the training of the 2D teacher network. As reported in Table \ref{tab:scanobjectnn-dgcnn}, our PointMCD still brings a significant performance boost from the original $82.8\%$ to the distilled $89.7\%$.

\begin{table}[t]
	\centering	
	\renewcommand\arraystretch{1.30}
	\setlength{\tabcolsep}{22.0pt}
	\caption{Performance of real-scanned object classification on ScanObjectNN (OBJ-BG) dataset using the DGCNN model, where the original and the distilled models are abbreviated as ``Orgn.'' and ``Dist.'', respectively.}
	\begin{tabular}{ c | c }
		\toprule[1.2pt]
		\textbf{DGCNN} & \textbf{ScanObjectNN} \\
		\hline
		\textit{\textbf{Orgn.}} & 82.8 \\
		\hline
		\textit{\textbf{Dist.}} & 89.7 \\
		\bottomrule[1.2pt]
	\end{tabular}
	\label{tab:scanobjectnn-dgcnn}
\end{table}

\subsubsection{Explorations of universal teacher knowledge} 

Performing cross-modal distillation for different downstream tasks using a single universal 2D teacher network is a highly interesting and promising direction. To verify this, we directly used an ImageNet-pretrained 2D CNN backbone encoder \cite{simonyan2014very}, which was also used in our preceding unsupervised learning experiments, to generate the teacher knowledge during the distillation process. As reported in Table \ref{tab:universal-teacher}, it turns out that using such a universal teacher network can still bring varying degrees of performance improvements, although the relative performance gains are relatively smaller compared with distilling with task-specific teacher networks.

\begin{table}[t]
	\centering
	\renewcommand\arraystretch{1.30}
	\setlength{\tabcolsep}{11.5pt}
	\caption{Performance comparison on multiple benchmark tasks, where the target student DGCNN model is distilled by the teacher knowledge exported from an ImageNet-pretrained 2D CNN image encoder \cite{simonyan2014very}.}
	\begin{tabular}{ c c | c c | c c }
		\toprule[1.2pt]
		\multicolumn{2}{c|}{\textbf{ModelNet40}} & \multicolumn{2}{c|}{\textbf{ScanObjectNN}} & \multicolumn{2}{c}{\textbf{ShapeNetPart}} \\
		\hline
		\multicolumn{1}{c|}{\textit{\textbf{Orgn.}}} & \textit{\textbf{Dist.}} & \multicolumn{1}{c|}{\textit{\textbf{Orgn.}}} & \textit{\textbf{Dist.}} & \multicolumn{1}{c|}{\textit{\textbf{Orgn.}}} & \textit{\textbf{Dist.}} \\
		\hline
		\multicolumn{1}{c|}{92.9} & 93.2 & \multicolumn{1}{c|}{82.8} & 84.4 & \multicolumn{1}{c|}{85.1} & 86.0 \\
		\bottomrule[1.2pt]
	\end{tabular}
	\label{tab:universal-teacher}
\end{table}

\section{Discussion} \label{sec:discussion}

\subsubsection{Design Philosophy} Throughout the whole methodology, our ultimate principle is to make the technical implementation of multi-view cross-modal distillation as concise as possible, while avoiding introducing complicated learning mechanisms in each core component and processing procedure. We believe that a simple but effective overall workflow will better validate the potential of multi-view cross-modal distillation. Therefore, we can reasonably expect that more advanced visual-geometric feature alignment/interaction techniques will further boost the current PointMCD learning framework.

\subsubsection{Experimental Evaluation} When evaluating the value of our method, we must particularly remind that major attention should be focused on whether PointMCD is able to contribute robust and stable performance gains when applied to different types of deep point encoders on diverse types of task scenarios. In other words, as PointMCD is claimed to serve as a universal plug-in component, one should observe whether the distilled students achieve better task-specific performance, instead of directly making comparisons with some other different learning frameworks. In the preceding experimental sections (e.g., Tables \ref{tab:modelnet40-classification}, \ref{tab:shapenetpart-segmentation}, and \ref{tab:modelnet40-unsupervised-learning}), we listed and compared different paradigms of methods only for demonstrating the significance of the actual performance gains brought by PointMCD.

In addition, during our development of experimental verifications, we noticed that previous works of multi-view learning only benchmarked over global geometry understanding tasks of either classification or retrieval, due to the inconvenience of preparing domain-specific datasets. This work makes efforts to extend additional application scenarios of fine-grained part segmentation and unsupervised learning, forming a much more comprehensive and persuasive evaluation protocol.

\subsubsection{In-depth Insights} Our encouraging experimental results indicate that, in addition to network design, we may as well pay attention to cross-modal data exploitation. In contrast to the availability of large-scale richly-annotated 2D visual data, collecting and labeling 3D geometric shapes can be much more inconvenient and expensive. Under such a context, the proposed image-to-point distillation approach is a promising and highly economical way of boosting 3D shape recognition performance.

\subsubsection{Promising Extensions} In terms of general categorization of knowledge distillation paradigms (as summarized in \cite{gou2021knowledge}), our proposed framework can be attributed to the most classic \textit{offline distillation}, which is composed of two separate stages, i.e., 1) pretrain the teacher model before distillation; 2) deploy the pretrained teacher model to guide the subsequent training of the student model. As we look towards the future, we are naturally motivated to investigate more diverse distillation paradigms, including \textit{online distillation} where the teacher model and the student model are simultaneously updated, as well as other data types \cite{carranza2022object} and richer application scenarios \cite{tang2022ydtr}.

Moreover, for another different application scenario of multiple modality fusion, we can also investigate multi-modal joint learning frameworks, where both visual and geometric signals are consumed as inputs in the inference phase, by introducing appropriate interaction mechanisms between image encoders and point cloud encoders.

Considering recent progress in visual-language joint modeling \cite{radford2021learning,li2022supervision}, it is also highly valuable to explore a unified representation learning framework simultaneously working on language, vision, and geometry signals.

\subsubsection{Limitations} 

Our current technical implementations may exhibit several typical aspects of potential limitations. First, in our viewpoint configuration scheme, all virtual cameras are positioned outside the underlying surface of 3D objects. For 3D shapes with severe self-occlusion or complex inner structures, the overall distillation process may not effectively impact the feature learning of all input points. Therefore, more adaptive viewpoint specification techniques are required to fully leverage the distillation power. Second, while our approach demonstrates satisfactory effectiveness in object-level task scenarios, adapting the same workflow to scene-level data domains, such as LiDAR point clouds with extremely sparse and non-uniform distributions, poses challenges due to difficulties in performing high-quality image rendering and accurate point-wise visibility checking. Consequently, more flexible feature alignment strategies for 2D image and 3D point cloud data modalities are expected to be further studied.

\section{Conclusion} \label{sec:conclusion}

We investigated the possibility of boosting deep 3D point cloud encoders by distilling discriminative cross-modal visual knowledge extracted from multi-view rendered images for 3D shape recognition task scenarios. Technically, we proposed a novel image-to-point distillation framework called PointMCD, serving as a universal plug-in component for generic deep set architectures. To facilitate view-specific heterogeneous feature alignment of paired visual and geometric descriptors, we also customized a simple but effective VAFP mechanism based on visibility checking. Extensive experiments and ablation studies strongly demonstrated the superiority and universality of our method. We believe that our work brings new insights in point cloud community and will motivate more explorations along this promising direction.



\qquad \\ \qquad \\ \qquad \\ \qquad \\ \qquad \\ \qquad \\ \qquad \\ \qquad \\

\qquad \\ \qquad \\ \qquad \\ \qquad \\ \qquad \\ \qquad \\ \qquad \\ \qquad \\

\begin{IEEEbiography}[{\includegraphics[width=1in,height=1.25in,clip,keepaspectratio]{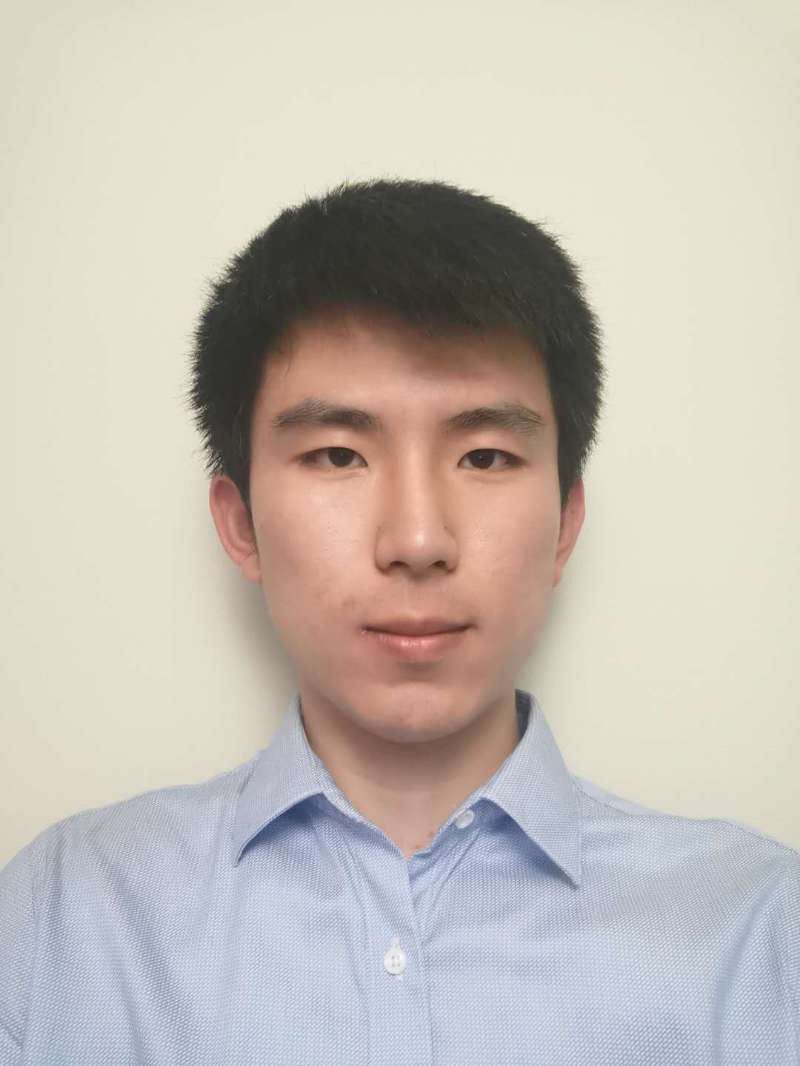}}]{Qijian Zhang}
	received the B.S. degree in Electronic Information Science and Technology from Beijing Normal University, Beijing, China, in 2019. Currently, he is a Ph.D. student with the Department of Computer Science, City University of Hong Kong, Hong Kong SAR. His research interests include 3D point cloud processing and geometric deep learning.
\end{IEEEbiography}

\begin{IEEEbiography}[{\includegraphics[width=1in,height=1.25in,clip,keepaspectratio]{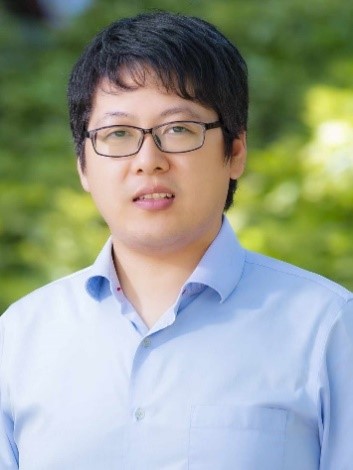}}]{Junhui Hou}
	 holds a B.Eng. degree in information engineering (Talented Students Program) from the South China University of Technology (SCUT), Guangzhou, China (2009), an M.Eng. degree in signal and information processing from Northwestern Polytechnical University (NPU), Xi’an, China (2012), and a Ph.D. degree from the School of Electrical and Electronic Engineering, Nanyang Technological University (NTU), Singapore (2016). He joined the Department of Computer Science at CityU as an Assistant Professor in 2017 and was promoted to Associate Professor in 2023. His research interests are multi-dimensional visual computing.
	
	Dr. Hou was the recipient of several prestigious awards, including the Chinese Government Award for Outstanding Students Study Abroad from China Scholarship Council in 2015 and the Early Career Award (3/381) from the Hong Kong Research Grants Council in 2018. He is an elected member of IEEE MSA-TC, VSPC-TC, and MMSP-TC. He is currently serving as an Associate Editor for \textit{IEEE Transactions on Circuits and Systems for Video Technology}, \textit{IEEE Transactions on Image Processing}, \textit{Signal Processing: Image Communication}, and \textit{The Visual Computer}. He also served as the Guest Editor for the \textit{IEEE Journal of Selected Topics in Applied Earth Observations and Remote Sensing} and \textit{Journal of Visual Communication and Image Representation} and an Area Chair of multiple conferences.
\end{IEEEbiography}

\begin{IEEEbiography}[{\includegraphics[width=1in,height=1.25in,clip,keepaspectratio]{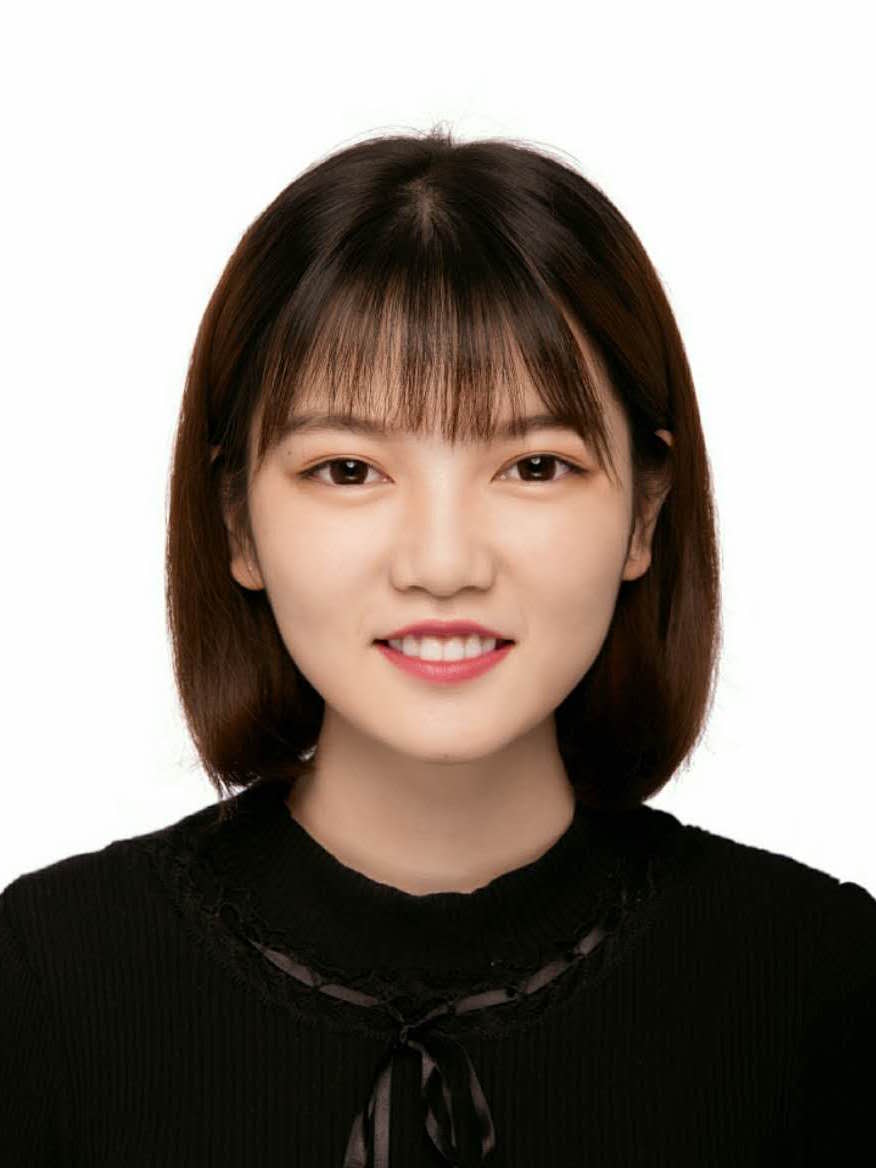}}]{Yue Qian}
	received the Ph.D. degree in Computer Science at the City University of Hong Kong, Hong Kong in 2022, and the B.S. and M.Phil. degrees in Mathematics from The Chinese University of Hong Kong, in 2014 and 2016, respectively. Her research interests include 3D point cloud and geometric processing.
\end{IEEEbiography}

\vfill

\end{document}